\documentclass[11pt,a4paper]{article}
\usepackage[hyperref]{emnlp2020}
\usepackage{times}
\usepackage{latexsym}
\usepackage{float}
\usepackage{xcolor}
\restylefloat{table}
\usepackage{amsmath}
\usepackage[utf8]{inputenc}
\usepackage{graphicx}
\DeclareMathOperator*{\argmax}{argmax}
\usepackage{adjustbox}
\usepackage{multirow}

\usepackage{microtype}

\aclfinalcopy % Uncomment this line for the final submission
\title{{S}ci{S}umm{P}ip: {A}n {U}nsupervised {S}cientific {P}aper {S}ummarization {P}ipeline }

\author{Jiaxin Ju$^1$, Ming Liu $^{2}$, Longxiang Gao$^2$, and Shirui Pan$^1$ \\
 $^1$Faculty of Information Technology, Monash University, Australia, VIC 3800\\
 $^2$School of Information Technology, Deakin University, Australia, VIC 3217 \\
 
 \texttt{jjuu0002@student.monash.edu}\\
  \texttt{\{m.liu, longxiang.gao\}@deakin.edu.au} \\
  \texttt{shirui.pan@monash.edu} \\
  }

\date{}

\begin{document}

\maketitle
\begin{abstract}
The Scholarly Document Processing (SDP) workshop is to encourage more efforts on natural language understanding of scientific task. It contains three shared tasks and we participate in the LongSumm shared task. In this paper, we describe our text summarization system, SciSummPip, inspired by SummPip \cite{zhao2020summpip} that is an unsupervised text summarization system for multi-document in news domain. Our SciSummPip includes a transformer-based language model SciBERT \cite{beltagy2019scibert} for contextual sentence representation, content selection with PageRank \cite{page1999pagerank}, sentence graph construction with both deep and linguistic information, sentence graph clustering and within-graph summary generation. Our work differs from previous method in that content selection and a summary length constraint is applied to adapt to the scientific domain. The experiment results on both training dataset and blind test dataset show the effectiveness of our method, and we empirically verify the robustness of modules used in SciSummPip with BERTScore \cite{zhang2019bertscore}.
\end{abstract}

\section{Introduction}

Text summarization aims at automatically generating a fluent and coherent summary that mainly contains the salient information from the source document(s). Two main categories are typically involved in the text summarization task, one is extractive approach \cite{luo2019reading,xu2019neural} which directly extracts salient sentences from the input text as the summary, and the other is abstractive approach \cite{sutskever2014sequence,see2017get,sharma2019entity} which imitates human behaviour to produce new sentences based on the extracted information from the given document.

In order to meet the requirements of modern data-driven methods, several large datasets have been presented. The majority of those datasets are for generic domain, but few available corpora from other task-specific domains. Most of existing state-the-art summarization systems \cite{liu2019text,zhou2020level,wang2020heterogeneous} target news or simple documents, and they are less adequate for summarizing scientific work due to the length and complexity. Those summarization systems cannot provide sufficient information conveyed in the scientific paper. 

The general domain have been paid enough attention, whereas the attention in scientific domain is far from enough. To address this point, the Scholarly Document Processing (SDP) workshop \cite{sdp2020} is held to accelerate scientific discovery in research community, they appeal to researchers for designing a summarization system that can generate a relatively long summary for scientific work.  

Since the release of Transformer \cite{vaswani2017attention} and BERT \cite{devlin2018bert}, much research has been carried out on involving them in their system. \citet{liu2019fine} modified the input sequence embedding and built several summarization-specific layers for extractive summarization. Similarly, \citet{liu2019text} present a novel document-level encoder based on BERT \cite{devlin2018bert} for both extractive summarization and abstractive summarization. In their model structure, the lower transformer represents adjacent sentences and the higher layer with self-attention mechanism represents the multi-sentence discourse. These works leverage the advantage of deep neural network, not taking into account the linguistic information. In contrast, \citet{zhao2020summpip}\footnote{https://github.com/mingzi151/SummPip} construct semantic clusters and sentence graphs for multi-document summarization, which involves linguistic information and discourse markers. In this paper, we followed the framework of \citet{zhao2020summpip} to construct our own unsupervised text summarization system. However, our model is different from the previous work: we modify the pipeline structure of multi-document summarization in the field of news to the single-document summarizer for summarizing scholarly documents, and we introduce two new steps to control the length of generated summary and to remove irrelevant sentences. 

Our contributions in this work can be summarized in the following aspects:

\begin{itemize}
\item  We highlight the importance of sentence embedding for scientific work. A variety of works focus on facilitating the process of obtaining sentence representation from a pre-trained language model on generic domain, while less attention is paid on other task-specific domains.
\item  We compare the performances between PageRank \cite{page1999pagerank} and the Maximal Marginal Relevance (MMR) \cite{carbonell1998use} in the content selection module. To our knowledge, no previous work compares their performances on scientific long document summarization task with deep neural representation.
\item  We experimentally verify that the effectiveness of the proposed model. We achieve better ROUGE results than original model on both training dataset and blind test dataset. Besides, our model is also evaluated on the BERTScore metric \cite{zhang2019bertscore} and the results indicate that our model is more robust to generate high quality summary.
\end{itemize}

\begin{table*}
\centering
\begin{tabular}{p{6.1cm}|c|c|c|c|c}
\hline
  \multirow{2}{*}{\textbf{Characteristics}}&\multicolumn{2}{c|}{\textbf{Extractive}}&\multicolumn{2}{c|}{\textbf{Abstractive}}&{\textbf{Test dataset}} \\ \cline{2-6}
                &Sci\_P&Ref\_S&Sci\_P&Ref\_S&Sci\_P \\ 
\hline
Range of corpus size (sentences)&[37,629]& [9,48]&[24,792]&[1,87]&[119,345] \\
\hline
Median value of corpus size (sentences)&186& 31&201&31&219\\
\hline
Range of sentence length (words) & [12,51] & [15,48]&[10,44]&[0.5,54]&[18,27] \\
\hline
Median value of sentence length (words) &26&27&26&21&22\\ 
\hline
\end{tabular}
\caption{Elementary data statistics for the LongSumm shared task of the Scholarly Document Processing @ EMNLP 2020. Sci\_P and Ref\_S represent scientific paper and reference summary, respectively.}
\end{table*}

\section{Related Work}

\paragraph{Text Summarization System} Most of recent text summarization systems leverage the advantages of deep neural networks, their encoder-decoder structures use either recurrent neural networks \cite{cheng2016neural,nallapati2016abstractive} or Transformer encoders \cite{zhang2019hibert,khandelwal2019sample}. Benefit of the sequence-to-sequence structure, a great progress in both extractive and abstractive document summarization is achieved. Though abstractive summarization has more potentials to generate interpretations in  a  human-like  fashion,  it  has  been  found  that sometimes repeatedly produces the same phrase or sentence \cite{suzuki2016cutting}, which greatly reduces the comprehensibility and readability. In contrast, extractive summarization performs better in fluency aspect and it can grammatical and accurately represent the source text. One potential issue in extractive summarization is that not all of information from the extracted sentence is important, which leads more redundancy in the generated summary. 

In the work of \citet{zhao2020summpip}, they apply graph structure and consider the discourse relationship between sentences rather than using encoder-decoder structure, and text compression is implemented in the final stage to reduce the redundancy in the generated sentences. However, their model is designed for multi-document summarization in the news domain, we extend their SummPip to single-document settings for scientific long articles.

\paragraph{Sentence Embedding Method} Term frequency–inverse document frequency (TF-IDF) is widely used in traditional NLP, but it cannot capture the semantic information and contextual relationship between sentences. Word2Vec \cite{mikolov2013distributed} is used in SummPip \cite{zhao2020summpip} to capture contextualized relationship, but this embedding method cannot solve the polysemous problem. More recently, BERT \cite{devlin2018bert} has achieved better performance in many NLP downstream tasks, but it is difficult to derive sentence embeddings. To solve this limitation, single sentences are passed to the BERT and two common ways to extract sentence representation are widely used: averaging the outputs and using the output of the [CLS] token \cite{may2019measuring,zhang2019bertscore}. 

\citet{xiao2018bertservice} develops a repository, bert-as-a-service\footnote{https://github.com/hanxiao/bert-as-service/}, which accelerates the process of extracting token and sentence embeddings from BERT \cite{devlin2018bert}. Lately, in order to find a better way to derive semantically similar sentence from language models, \citet{reimers2019sentence} present SBERT. However, above works help facilitate workload in generic domain rather than task-specific domain. 

\paragraph{Content Selection} Graph is an intuitive structure for utilizing the relation information between sentences. Some work \cite{mihalcea2004textrank,erkan2004lexrank}  focuses on selecting salient sentences by leveraging graph-based ranking methods. Inspired by PageRank algorithm \cite{page1999pagerank}, they consider the document as a graph where sentences are vertices and edges represent the relations between two sentences.  Shortly thereafter, some researchers \cite{carbonell1998use,kurmi2014text,mao2020multi} involved a query-biased strategy, the Maximal Marginal Relevance (MMR) \cite{carbonell1998use}, in their summarizers. MMR tries to balance the relevance and diversity by controlling the trade-off parameter $\lambda$. The first part of the formula controls query relevance and the second part controls diversity. 
\begin{align*}
MMR &= \argmax_{S_i \in \mathcal{C}} \lambda Sim_1(S_i,Q) \\ 
& - (1- \lambda ) \argmax_{S_j \in \mathcal{S}} Sim_2(S_i, S_j)
\end{align*}
Where $C$ is the set of candidate sentences, $S$ is the set of extracted sentences, $Q$ is the query embedding, $S_i, S_j$ are sentence embeddings of candidate sentences $i$ and $j$, respectively. $Sim$ indicates the cosine similarity between two embeddings.

Though this approach have been proved that it outperforms generic summarization approaches in the information retrieval task, to our knowledge, there is no previous work compared it with PageRank algorithm on scientific long document summarization task. Our work incorporates deep neural representations into both PageRank algorithm and MMR strategy and shows the comparison between these two methods in the field of scientific work for both extractive and abstractive summarization.

\section{Dataset Pre-processing}
The training dataset provided by the LongSumm shared task consists of 2236 scientific papers, of which 1705 are for extractive method and 531 are for abstractive method. The reference extractive summaries are generated by TalkSumm \cite{lev2019talksumm} that extracts sentences appeared in associated conference videos, while the abstractive summaries are collected from blogs written by researchers.

\paragraph{Download paper} We download the training corpus from the given URLs (for abstractive) and the script (for extractive).

\paragraph{Paper Parsing} All of papers are parsed from PDF form into JSON structure by using Science-Parse\footnote{https://github.com/allenai/science-parse}. It outputs a JSON file for each PDF, which contains the title, abstract text, metadata, and the text of each section in the paper. 

\paragraph{Text processing} We concatenate each section text as the paper text. Then sentences are segmented by using the NLTK library, and each sentence is tokenized as well. Table 1 reports the result of the statistics analysis for both training dataset and test dataset, and we can see that the number of sentences in some reference summaries is far less than required length of generated summary, 600 words, which may lead a bias in the evaluation.

\section{System overview}

We adopt the SummPip \cite{zhao2020summpip} as our baseline model, and we modify the pipeline architecture for summarizing scholarly documents. Two new steps are introduce for adapting scientific domain, one is to remove irrelevant sentences and the other is to control the length of generated summary. In the following subsections, we will specify each component in the SciSummPip.

\subsection{Embedding Method}
\paragraph{Pretrained language model} In this paper, we apply a publicly available large-scale language model, SciBERT \cite{beltagy2019scibert}, which is pretrained based on BERT \cite{devlin2018bert} and extends the idea of word embeddings by learning contextual representations from large-scale scientific corpora. This is implemented in Pytorch using Transformers established by \citet{Wolf2019HuggingFacesTS}\footnote{https://github.com/huggingface/transformers}.

\paragraph{Sentence embedding} Using more accurate sentence embeddings can improve the performance of summarization system in language understanding. In SciSummPip, we average the output of SciBERT from the second layer to the last layer. In addition, we also experiment with other embedding methods and the the results show that this is a more accurate way to represent scientific sentences.

\subsection{Sentence Graph Construction}
\paragraph{Content selection} Not all of sentences should be involved in the summary, so we include content selection step before constructing sentence graph. We build a matrix to store the similarity between each two sentences, then PageRank \cite{page1999pagerank} algorithm is implemented to rank all of sentences. Sentences with lower score will be deleted from the candidate list, here we introduce a new step to control the ratio of removed sentences.

\paragraph{Graph construction} We construct the sentence graph, where each node represents a sentence, and nodes are connected if they meet the linguistic requirements. To identify this structure, we borrow the components from the previous work \cite{zhao2020summpip}. Specifically, this pipeline consists of discovering deverbal noun reference, finding the same entity continuation, recognizing discourse markers, and calculating sentence similarity by taking the cosine similarity. 

\subsection{Text Generation}
\paragraph{Spectral clustering} After identifying pairwise sentence connection, we involve a new step for determining the number of clusters. This is to control the length of generated summary so that the summary varies with the length of the original paper.
\paragraph{Multi-sentences compression} This module \cite{boudin2013keyphrase} is to generate a single summary sentence from each sentence cluster. Sentences with similar semantic information will be compressed by building a word graph. Considering the key phrases and discourse structure, so that the reconstructed sentence will have higher score. Select the sentence with the highest score as the summary sentence, and then combine all reconstructed summary sentences as the generated summary.

\begin{table*}
\centering
\begin{tabular}{p{3.3cm}| p{1.3cm} p{1.3cm} p{1.3cm} p{1.3cm} p{1.3cm} p{1.3cm}}
 \hline
    & R1\_F & R1\_R & R2\_F & R2\_R &RL\_F & RL\_R\\ [0.5ex] 
 \hline
 \textbf{Extractive dataset} \\
 \hline
 Scibert-summarizer &\textbf{58.13} & \textbf{57.53}&\textbf{27.20}&\textbf{26.82}&\textbf{28.65} & \textbf{28.29} \\ 
 TextRank &57.42&57.31&26.38& 26.48&28.48&27.59 \\ 
 LextRank  &46.23&36.38&20.71&16.33&21.34&16.76 \\ 
 MMR\textsubscript{$Sci$} ($\lambda$=0.5) &55.24&55.48&23.74&23.85&21.00&21.11\\
 \hline
 \textbf{Abstractive dataset} \\
 \hline
 SummPip\textsubscript{$+PR$} &36.17&32.73&8.36&7.27&14.80&13.83 \\ 
 SciSummPip\textsubscript{$PR$} &\textbf{40.90}&\textbf{43.09}&\textbf{9.52}&\textbf{9.83}&\textbf{15.47}&\textbf{17.26}\\
 SciSummPip\textsubscript{$MMR0.2$} &32.34&28.31&6.54&5.48&13.60&12.37\\
 SciSummPip\textsubscript{$MMR0.5$} &30.69&25.63&6.64&5.32&13.37&11.56\\
 SciSummPip\textsubscript{$MMR0.8$} &33.06&27.88&7.58&6.17&14.18&12.39\\
 \hline
 \textbf{Blind Test Dataset} \\
 \hline
 Scibert-summarizer &\textbf{49.16} &\textbf{49.35}&\textbf{12.80}&\textbf{12.76}&\textbf{18.31}&\textbf{18.33} \\ 
  SciSummPip &47.37&40.89&13.35&11.40&17.54&15.02\\
 SummPip &38.62&30.16&9.01&6.95&15.15&11.74\\[1ex]
 \hline
\end{tabular}
\caption{ROUGE scores reported on the training dataset and the blind test dataset. Best results are in \textbf{boldface}. The reference extractive summary and abstractive summary are generated by TalkSumm \cite{lev2019talksumm} and collected from online blogs, respectively. MMR\textsubscript{$Sci$} indicates we implement MMR algorithm with sentence embeddings derived from SciBERT\cite{beltagy2019scibert}. SciSummPip\textsubscript{$PR$} and SciSummPip\textsubscript{$MMR$} are our model with different content selection modules, and the number follow the MMR is the setting for trade-off parameter $\lambda$. As SummPip cannot effectively run on large scale corpora of long document, we add content selection module and shown as SummPip\textsubscript{$+PR$}.}
\end{table*}

\section{Experiment Setup}
\subsection{Implementation Details}
\paragraph{Extractive summarization Task} We use SciBERT for sentence embedding in our pipeline, so for extractive text summarization task we directly use Scibert-summarizer\footnote{bert-extractive-summarizer: https://pypi.org/project/bert-extractive-summarizer/} with the fixed length range (from 60 to 600 words). 

\paragraph{Abstractive summarization Task} We implement our pipeline, SciSummPip, in abstractive summarization task, and we compare the performances of PageRank algorithm and of MMR strategy in the content selection module. For PageRank algorithm, we set a cutoff ratio that is a new introduced parameter for removing irrelevant sentences andthe empirical results show that setting it as 0.25 achieves better performance. For the MMR strategy, we set 0.2, 0.5, 0.8 for the trade-off parameter in the experiment, respectively. To control the generated summary length, we introduce another new parameter, extended ratio, to modify the number of clusters based on the number of ranking sentences. In our pipeline,we set it as 0.3.

\subsection{Comparison Systems}
For extractive task, we compare our model with the following unsupervised summarization models:

\paragraph{TextRank \cite{barrios2016variations}} TextRank \cite{mihalcea2004textrank} applies a variation of PageRank algorithm \cite{page1999pagerank} over a graph-based structure, and it produces a list of ranked elements in the graph without the need of a training corpus. TextRank implemented in this paper is produced by \citet{barrios2016variations}, they change the similarity function to Okapi BM25 so that the performance is better than the original textRank model. We set the output summary with the fixed length 600 words.

\paragraph{LexRank \cite{erkan2004lexrank}} Similar with textRank \cite{mihalcea2004textrank}, LexRank also applies PageRank algorithm and leverages a graph structure for summarization. Differently, textRank calculate the similarity based on the number of words two sentences have in common, while LexRank uses cosine similarity of TF-IDF vectors. 

\paragraph{MMR \cite{carbonell1998use}} MMR is a query-biased summarization approach, it tries to balance the relevance and diversity by controlling the trade-off parameter $\lambda$. In the previous works, the similarity usually calculate based on TF-IDF, but in our implementation we use sentence embeddings derived from the output of SciBERT \cite{beltagy2019scibert}. In addition, we set the document title as the query and the fixed length of generated summary is set as 600 words.

For abstractive task, we apply different sentence embedding methods in SciSummPip: 
\begin{itemize}
\item SciBERT \cite{beltagy2019scibert}: We implement two common strategies for sentence embeddings derived from SciBERT model: averaging the output from the second to the last layer and using [CLS] token embedding.
\item SummPip \cite{zhao2020summpip}: We use the same embedding method with the original pipeline to compare the performance.
\item SBERT \cite{reimers2019sentence}: This is a modification of the BERT network using siamese and triplet networks in order to find semantically similar sentences in vector space. Their empirical results indicate that their method is better than those two common embedding strategies, so we incorporate it into SciSummPip as a comparison.
\end{itemize}

\section{Evaluation and Results}
\subsection{Experiment result on training dataset}
\paragraph{Extractive summaries} The training dataset for extractive method consists of 1705 papers, of which one paper cannot be parsed. Thus, we evaluate 1704 papers with the ROUGE metric\cite{lin2003automatic} in our experiments.

As displayed in Table 2, the Scibert-summarizer achieves better ROUGE scores than all other compared systems. We implement MMR algorithm with sentence embedding derived from averaging SciBERT \cite{beltagy2019scibert} output, and we can see it performs better than LexRank \cite{erkan2004lexrank} but worse than the textRank model \cite{barrios2016variations} with the Okapi BM25 similarity function. Therefore, we can verify that PageRank ranking algorithm performers better than MMR strategy in extractive task.

\paragraph{Abstractive summaries} For abstractive experiments, we collect 530 summaries in total as one paper cannot be parsed by Science-parse. 

We implement SciSummPip with different parameter settings to find out the best one. The number of words in each sentence is set from 15 to 29, then we observe that the summary with 26 words in each sentence achieves the best performance. We incorporate PageRank algorithm \cite{page1999pagerank} and MMR algorithm \cite{carbonell1998use} into SciSummPip content selection module, respectively. As displayed in Table 2, it is not surprising to see SciSummPip with PageRank algorithm outperforms all of settings for SciSummPip with MMR algorithm, because the performance of textRank is better than that of MMR in the extractive task.

\begin{table}
\centering
\begin{tabular}{p{4.15cm} p{0.61cm} p{0.61cm} p{0.61cm}}
 \hline
 \textbf{Sentence Embedding}& R1\_F & R2\_F &RL\_F\\
 \hline
 Avg. SciBERT embeddings &40.90&9.52&15.47\\
 Special token embedding &39.27&8.81&15.09\\
 Word2Vec &36.17&8.36&14.80\\
 SBERT &39.75&9.41&15.27\\
 \hline
\end{tabular}
\caption{ROUGE F1 scores for SciSummPip with different sentence embedding methods. Special token embedding method is extracting [CLS] token embedding from SciBERT \cite{beltagy2019scibert} output. }
\end{table}

\begin{table}
\centering
\begin{tabular}{p{4.15cm} p{0.61cm} p{0.61cm} p{0.61cm}}
 \hline
 \textbf{Sentence Embedding}& R1\_R & R2\_R &RL\_R\\
 \hline
 Avg. SciBERT embeddings &43.09&9.83&17.26\\
 Special token embedding &39.99&8.75&16.13\\
 Word2Vec &32.73&7.27&13.83\\
 SBERT &41.53&9.56&16.73\\
 \hline
\end{tabular}
\caption{ROUGE Recall results for SciSummPip with different sentence embedding methods.}
\end{table}

\subsection{Experiment result on test dataset}
The blind test dataset consists of 22 scientific papers\footnote{Test dataset: https://github.com/guyfe/LongSumm}. It does not declare the blind test data is for extractive summarizer or abstractive summarizer, so we implement both Scibert-summarizer and SciSummPip on it.  Comparing with the SummPip \cite{zhao2020summpip}, the experiment results verify that our new pipeline architecture significantly improve the performance. In addition, we try different number of words generated in each sentence and we find that setting it closes to the median value of that in scientific papers would gain higher score. Besides, although extractive model gains the highest ROUGE score, we still can see our SciSummPip is competitive.

\begin{table}
\centering
\begin{tabular}{p{2.45cm} p{1.1cm} p{1.0cm} p{1.45cm}}
 \hline
   & Precision & Recall &F1-Score\\
 \hline
 \hline
 SciSummPip &\textbf{0.807}&0.800&\textbf{0.815}\\
 SciSummPip\textsubscript{$MMR$} &0.806&0.810&0.808 \\
 SummPip\textsubscript{$+PR$} &0.794&0.813&0.806\\
 SBERT &0.795&\textbf{0.814}&0.804\\
 \hline
 \hline
\end{tabular}
\caption{BERTScore reported on abstractive training dataset to investigate text generation ability of our model. SBERT means we use use SBERT sentence embedding method in SciSummPip.}
\end{table}

\subsection{Different Sentence Embedding Methods}
To find out a more accurate method for representing scientific sentences, we incorporate different embedding strategies into SciSummPip. Performances reported in Table 3 and Table 4 indicate that our model ranks highest with averaging the output of SciBERT \cite{beltagy2019scibert} method. SBERT \cite{reimers2019sentence} shows competitive performance even though it is designed for generic domain. In fact, utilizing SBERT significantly reduce the workload of extracting sentence embedding, but it is not sufficient enough for representing scientific sentence.

\begin{figure*}[!t]
    \centering
       \includegraphics[width=1.0\textwidth]{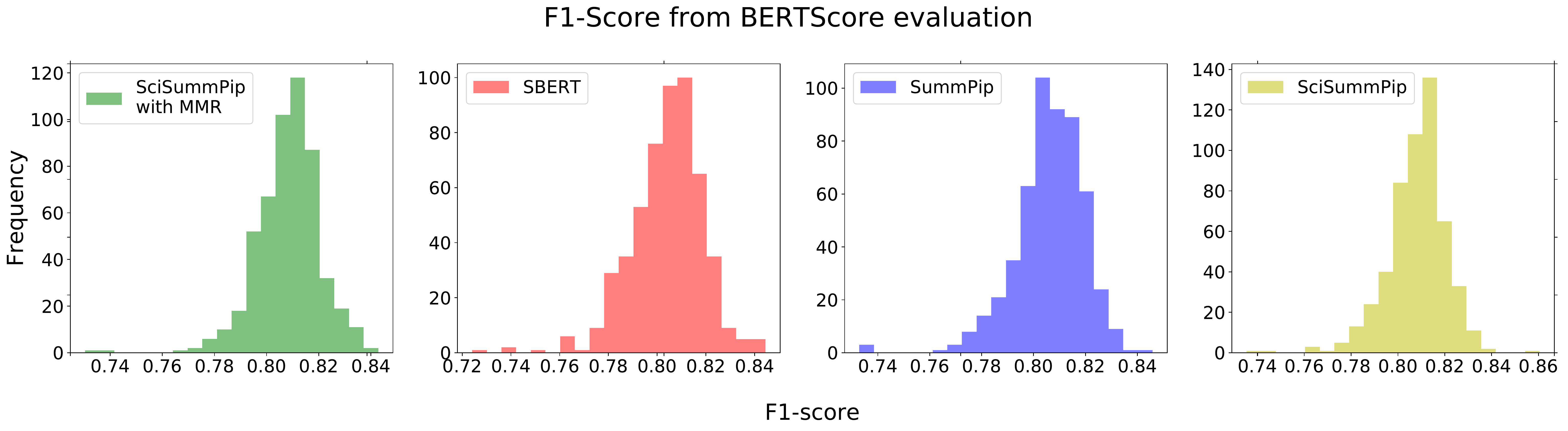}\\
       \caption{The histogram distribution of F1-score evaluated by BERTScore metric for each model reported in Table 5. X-axis indicates data range of F1-score and Y-axis indicates the frequency of the data in each bin. In order to ensure the bin data range for each distribution is same, we set the data range of each bin as 0.005 so that the parameter, bins, is set as $int(data\ range\ of\ F1-score/0.005)$.}
\end{figure*}

\subsection{BERTScore Evaluation}
We evaluate models on BERTScore \cite{zhang2019bertscore}, an automatic evaluation metric for text generation, to investigate the ability of writing abstractive summary. BERTScore calculates a similarity score for each token in the candidate sentence with each token in the reference sentence by leveraging  contextual embeddings. As can be seen in Table 5,  SciSummPip achieves highest precision and F1-score while SBERT gains the highest recall. This proves that the summary generated by our model is more informative and representative. Since BERTScore utilizes Bert \cite{devlin2018bert} to calculate similarity score, the max length of input sequence is 512 tokens, which limits the performance of relatively long summary. 

We further investigate the distribution of F1-score from BERTScore evaluation. As shown in figure 1, although these models achieve similar performance, the F1-score distribution of SciSummPip obviously more stable than others. SciSummPip achieve the highest frequency in the range of 0.80-0.82, which means near 140 generated summaries gain around 0.81 F1-score. Therefore, we can say that our model is more robust for summarizing scientific work in abstractive task.

\begin{table}[h!]
\small
\begin{tabular}{|p{7.25cm}|}
\hline \vspace{0.02cm}
\textbf{Extractive Reference Summary:}\\
\textcolor{blue}{The analysis of emotions in texts is an important task in NLP. }Traditional studies treat this task as a pipeline of two separated sub-tasks: emotion classification and emotion cause detection. The former identifies the category of an emotion and the latter detects the cause of an emotion. This separated framework makes each sub-task more flexible to deal with, but it neglects the relevance between the two sub-tasks. In this paper, we use the human-labeled emotion corpus provided by Cheng et al. (2017) as our experimental data (namely Cheng emotion corpus). \textcolor{blue}{Cheng emotion corpus can be considered as a collection of subtweets.} For each emotion in a subtweet, all emotion keywords expressing the emotion are selected, and then the class and the cause of the emotion are annotated. (...) \\[0.5ex]
 \hline\vspace{0.02cm}
\textbf{Scibert-summarizer:} \\
\textcolor{blue}{The analysis of emotions in texts is an important task in NLP.} \textcolor{blue}{Cheng emotion corpus can be considered as a collection of subtweets.} Given an instance which is a pair of \textless an emotion keyword, a clause in the subtweet\textgreater, ECause assigns a binary label to the instance to indicates the presence of a causal relation. The input text of an ECause instance also has three sequences of words: the emotion keyword (i.e. EmoKW), the current clause (i.e. CauseCL) and the context between EmoKW and CauseCL. The BiLSTM layer focuses on the extraction of sequence features, and the attention layer focuses on the learning of word importance (weights). (...)  \\[0.5ex]
\hline
\end{tabular}
\caption{Example of the generated extractive summary compared with reference summary that is generated by TalkSumm \cite{lev2019talksumm}. Text in the same color indicates the content they describe is the same. Due to the length constraint, we omit part of the generated summary and shown as (...).}
\end{table}

\begin{table}[h!]
\small
\begin{tabular}{|p{7.25cm}|}
\hline \vspace{0.02cm}
\textbf{Abstractive Reference Summary:}\\
\textcolor{red}{The paper proposes a two-stage synthesis network that can perform transfer learning for the task of machine comprehension.} The problem is the following: \textcolor{blue}{We have a domain DS for which we have labelled dataset of question-answer pairs and another domain DT for which we do not have any labelled dataset.} We use the data for domain DS to train SynNet and use that to generate synthetic question-answer pairs for domain DT. Now we can train a machine comprehension model M on DS and finetune using the synthetic data for DT. SynNet Works in two stages:  Answer Synthesis - Given a text paragraph, generate an answer. (...) After the word vector, append a ‘1’ if the word was part of the candidate answer else append a ‘0’. \textcolor{magenta}{Feed to a Bi-LSTM network (encoder-decoder) where the decoder conditions on the representation generated by the encoder as well as the question tokens generated so far.} (...) \\[0.5ex]
\hline \vspace{0.02cm}
\textbf{SciSummPip:} \\
the ability to quickly use a mc model \textcolor{blue}{trained on one domain to answer questions over paragraphs from another with no annotated data.} recent work generated synthetic data generated questions leads to improved performance, we use a model where the answer synthesis and question types. we generate the answer first because answers are usually key semantic concepts, while questions can transfer a mc model trained on another domain. \textcolor{red}{when we ensemble a bidaf model fs we use the two-stage synnet to generate data tuples to directly boost performance boost.} (...) however, unlike machine translation , for tasks like mc, we need to synthesize both the question and answers given the context paragraph. (...) \textcolor{magenta}{the first stage of the model, an answer synthesis module , uses a Bi-directional LSTM to predict iob tags on the input paragraph}, which mark out key semantic concepts that are likely answers.(...)\\[0.5ex]
\hline
\end{tabular}
\caption{Example of the generated abstractive summary compared with reference summary that is collected from researcher's blog. Text in the same color indicates the content they describe is similar. Due to the length constraint, we omit part of the generated summary and shown as (...).}
\end{table}

\subsection{Human Analysis}
We further manually inspect the generated summary to explore if our model can capture the salient information from given document. Table 6 and Table 7 display an example of generated summary compared with the corresponding reference summary in the training dataset. The abstractive reference summary is collected from the online blog written by the researcher, so it is more difficult to capture the similar description in the generated summary. However, As shown in table 7, our model successfully write some similar context in the final output. Notwithstanding, we have to say the readability and grammatically of the generated summary still need to be improved.

For blind test dataset, we also inspect the extractive summary and abstractive summary for the same paper. We find that the Scibert-summarizer tends to extract the sentence appeared in the early part of the paper, and the generated summary usually lack of logicality and consistency. In contrast, the summary produced by SciSummPip is more logical and contains more salient information about the methodology and the experiment. Although Scibert-summarizer gains higher ROUGE score on the blind test dataset, the summary generated by our model is more consistent with the purpose of the LongSumm Shared Task.

\section{Conclusion and Limitation}
In this paper, we have presented the modified unsupervised pipeline architecture, SciSummPip,  that leverages a transformer-based language model for summarizing scientific papers. We add content selection module and two steps to remove irrelevant sentences and to control the length of generated summary. After that, the linguistic knowledge will be incorporated into the process of multi-sentences compression for summarizing scientific work. The experiment results of automatic evaluation prove that our new pipeline significantly improves the overall performance on both training and blind test dataset. Besides, through manual inspection we find that our model indeed capture the salient information from the given source document. However, we have to admit that the readability of generated summary needs to be improved.

We incorporated deep neural representation into both MMR \cite{carbonell1998use} strategy and PageRank \cite{page1999pagerank} algorithm. Even though MMR strategy performs better in information retrieval task, we empirically verified that it is not sufficient for our model to summarize scientific work. MMR is a query-biased approach and we chose the title as query in our implementation, thus the potential reason for worse performance may be the query we chose is not effective enough.

To investigate a sentence embedding method for sufficiently summarizing scholarly document, we compared the performances among several embedding strategies and we also evaluated their performances on both ROUGE metric and BERTScore metric. Although averaging the output of SciBERT \cite{beltagy2019scibert} achieves better performance, the workload of using it to extract sentence embeddings is heavier than that of directly using SBERT \cite{reimers2019sentence}. There is enough work for generic domain while the attention paid for task-specific domain is far from enough, therefore we appeal to researchers for making more efforts on task-specific domain in their further research.

\section{Future work}
As the future, we will evaluate our pipeline on larger scientific datasets to show the effectiveness and robustness, and we also would like to conduct a analysis on the faithfulness and the level of abstraction for the generated summary.

\section*{Acknowledgments}
We would like to thank the anonymous reviewer(s) for helpful comments and suggestions.

\bibliographystyle{acl_natbib.bst}
\bibliography{reference}

\end{document}